\documentclass[paper=a4, fontsize=11pt]{scrartcl}
\usepackage[utf8x]{inputenc} 
\usepackage{csquotes}
\usepackage{amsmath,amsfonts,amsthm} 
\usepackage[pdftex]{graphicx}	

\usepackage{multirow, graphicx}

\usepackage[backend=biber]{biblatex}
\usepackage{algorithm}
\usepackage{algpseudocode}
\usepackage{hyperref}
\usepackage[table]{xcolor}
\usepackage[dvipsnames]{xcolor}
\usepackage{tabularx}
\colorlet{LightLavender}{Lavender!40!}
\usepackage{tcolorbox}

\numberwithin{equation}{section}		
\numberwithin{figure}{section}			
\numberwithin{table}{section}				

\usepackage{array}
\renewcommand{\arraystretch}{1.2}

\title{Learner-based Concept Drift Detection: Analysis and Evaluation}
\author{Md Moman Ul Haque Khan and Samira Sadaoui}
\publishers{
Department of Computer Science, University of Regina, Canada\\
\texttt{MdMomanUlHaque.Khan@uregina.ca, Samira.Sadaoui@uregina.ca}
}

\date{}

\begin{document}
\maketitle
\tableofcontents 
\vspace*{1.5cm}

\noindent \textbf{Abstract: }
Machine learning algorithms deployed for evolving streaming environments must handle the non-stationary data distributions, commonly referred to as concept drift. The presence of concept drift poses a major challenge for many real-world applications because it can severely degrade their predictive performance, hindering their ability to support robust decision-making. Consequently, the timely and efficient detection of drift events is critical for sustaining high accuracy over time. This study examines theoretically the concept drift characteristics and numerous drift detection algorithms across several categories. Furthermore, we evaluate their performance on both synthetic and real-world datasets exhibiting diverse streaming scenarios and drift characteristics, such as abrupt and gradual changes. This study aims to enhance understanding of the complex notion of concept drift characteristics and behavior of drift detectors, along with their applicability to diverse contexts. \\

\textbf{Keywords: } Concept drift types, Transition speeds, Sudden and gradual drifts, Concept drift detection, Learner-based detection, SPC methods, Windowing methods, Ensemble methods, Synthetic stream datasets, Implementation frameworks. 

\section{Introduction}
Nowadays, most applications evolve over time, such as fraud detection, network intrusion, health monitoring, financial markets, predictive maintenance and environment monitoring. For example, in e-payment systems, fraudsters often devise new strategies to manipulate the system's vulnerabilities, and legitimate buyers adopt new payment methods (e.g., using new mobile wallets) \cite{Malekian2013, Bayram2020}. Regarding the environment monitoring systems, sensors can result in shifts due to new pollution causes \cite{Zhang2021, Mehmood2021}. Health monitoring devices may also encounter drift because a patient's lifestyle, age and health status (health is improving or deteriorating) change over time \cite{Lu2020b}. Industrial equipment also experiences drift as it ages and degrades, and monitoring models must detect failures early \cite{Togbe2021, Nunes2024}. These non-stationary applications process an influx of data produced non-stop, often quickly, and where data can shift unpredictably. These changes lead to a divergence between training and operational data-distributions over time \cite{iwashita2019, khamassi2018}. Such changes in the class distribution and/or feature distribution, known as concept drift \cite{hu2019}, often degrade the performance of predictive models \cite{hu2019}. Indeed, ML models trained on old data patterns become invalid and obsolete, leading to poor performance and false alarms. 

Concept drift occurs when the statistical properties of the incoming data change or the relationships between the independent variables (features) and target variable (class label) change. The evolving data-generating environments present memory and computational challenges, making it critical to develop adaptive algorithms that can effectively address them \cite{khamassi2018}: 1) Data arrives at high speed, necessitating algorithms with fast data and drift analyzing capabilities, and 2) Data are continuous, making it impractical to store all the data, which requires algorithms to process each sample individually, or select the most relevant data or summarize data on the fly. For each specific application, the model developers must anticipate possible changes in future data and choose the most effective strategies to adapt ML models to real-time changes in data properties, such as discarding outdated knowledge and retraining the models on the adjusted feature space. \\

Our study provides a comprehensive survey of concept drift within the supervised learning setting, with a particular focus on data stream classification. To support a clearer understanding of the complex nature of concept drift, we first examine its key characteristics through several concrete examples, including different drift types, such as real and virtual drift, as well as different rates of change, such as abrupt, gradual, incremental and recurring drift. The focus is placed on learner-based drift detection methods, which are the most widely studied and applied approaches. These methods typically monitor the behavior or performance of a learning model over time in order to identify significant changes that may indicate the presence of drift. We outline a general algorithmic framework for learner-based detection and examine its main categories: Statistical Control Process (SPC), Window-based and Ensemble-based. For each category, we review a selection of representative drift detectors (with a tally of 15 methods). 

In addition to the survey part, we conduct a thorough empirical evaluation of the reviewed methods on several synthetic (under gradual and sudden drifts) and real-world data streams. We choose artificial datasets because the drift locations are known a apriori, which allows for a more precise assessment of each method’s ability to detect both sudden and gradual drifts. We also consider real-world datasets to evaluate the practical behavior of the methods under realistic conditions, where drifts may be noisy or difficult to identify. By combining a robust review with an extensive experimental evaluation and comparison, our study aims to provide researchers with a clearer and more practical understanding of learner-based detection methods. \\

The remainder of the paper is structured as follows. Section 2 presents the formal definitions of concept drift and discusses its main characteristics using real-world examples. Section 3 examines different classifications of drift detection methods, including active vs. passive, online vs. block, learner-based vs. distribution-based,  and SPC-based vs. window-based vs. ensemble-based. Sections 4, 5 and 6 present the theoretical foundations of SPC-, window- and ensemble-based detectors, respectively. Section 7 describes the existing implementation frameworks for these various methods and also introduces the stream datasets used in the experimental evaluation. Sections 8, 9, 10 and 11 evaluate and compare the performance of SPC-, window- and ensemble-based methods. Section 12 concludes our work.

\section{Characteristics of Concept Drift}
\subsection{Formal Definitions}
Concept drift occurs when learning from dynamic data streams, defined as continuous and limitless sequence of samples, $[(X_0, y_0), ...., (X_i, y_i)]$, where $X$ is a multi-dimensional feature vector with a corresponding label $y$ \cite{hoens2012}. The Bayes' decision theorem allows to compute the probability that $X$ is an instance of class $y$ \cite{gama2014}: \[P(y|X) = \frac{P(X|y)P(y)}{P(X)}\]

Here,  $P(X|y)$ denotes the likelihood of the input features given the target label, $P(y)$ is the prior probability distribution of the output variable, and $P(X)$ is the unconditional probability distribution of the input features. In predictive modeling, a concept refers to the specific joint probability of input features and target class, $P(X,y)$, which encompasses the prior class probability $P(y)$ and class-conditional probability $P(X|y)$ as follows \cite{bayram2022, khamassi2018}: $P(X, y) = P(y)P(X|y)$
\medskip 

Concept drift can be defined as a change in the joint probability distribution between two distinct time steps $t$ and $t+w$. Here, $t$ represents a specific point or interval in time, and $w$ denotes the window over which the change in data distribution has been identified \cite{bayram2022, khamassi2018}. So, concept drift occurs when $P_{t}(X,y) \neq P_{t+w}(X,y)$ \cite{bayram2022, khamassi2018}. 
\subsection{Types of Drifts} 
The types of drift are categorized based on the aspects of the probability distribution that is changing. 
\medskip 

\textbf{A. Real Drift:} Known as actual or true drift, this type occurs when there a change in the posterior probability $P(y|X)$, which can affect the decision boundary of ML models. This drift can be a result of changes in the class distribution $P(y)$ and likelihood distribution $P(X|y)$. It indicates a fundamental change in the concept the model is trying to predict, which over time renders the model less effective \cite{bayram2022,khamassi2018}: $P_{t}(y|X) \neq P_{t+w}(y|X)$.
\medskip 

\textbf{An example:} We assume we have a model that predicts the likelihood of an illness based on the symptoms observed ($X$). In this case, real drift refers to changes in the conditional probability of the illnesses given the symptoms, i.e. $P(Disease|X)$. For instance, the changes are due to an update in medical knowledge or emergence of new illness strains. \\ 

\textbf{B. Virtual Drift:} Known as covariate drift, this type refers to changes in the feature-value distribution $P(X)$, without altering the relationship between the input features and the label i.e., $P(y|X)$. With this type, the model might encounter samples that are underrepresented or differ  from those in the training dataset, leading to accuracy issues \cite{bayram2022,khamassi2018}:  $P_{t}(X) \neq P_{t+w}(X)$.
\medskip 

\textbf{An example:} In predicting weather conditions, suppose the model was trained with data from the autumn season. If the model starts receiving data from winter  (without any change in $P(Weather|X)$), this would be an instance of virtual drift. \\

\textbf{C. Mixed Drift:} Both real and virtual drifts can happen simultaneously in real-world scenarios, involving changes in both the prior probability of classes and probability of features \cite{zliobaite2010, xiang2023}: $P_{t}(y|X) \neq P_{t+w}(y|X)~~and~~ P_{t}(X) \neq P_{t+w}(X)$. 
\medskip 

\textbf{An example}: In credit card fraud detection, we usually encounter real and virtual shifts, such as the consumer's spending habits change during the holiday season (real drift) and fraudsters change their tactic strategies (virtual drift). 
\subsection{Transitions of Drifts}
Transitions represent the speed of change from the old concept to the new one. They can be quantified by the number of samples over which the drift occurs until the new concept is established. These transitions are essential for understanding how to adapt the decision models to maintain accuracy as data evolves. There are four transition types \cite{gama2014, lu2020a}, and for each type, concrete examples are provided based on studies such as \cite{bayram2022, Malekian2013, lu2020a}. 

\begin{itemize}
\item \textbf{A. Sudden Drift:} Occurs when the data distribution changes abruptly from the old concept to a new one at a precise timestamp, which degrades instantly the decision models. Abrupt drifts that can drastically impact the learned models. 

\textbf{Examples:}  1) A sudden change in consumer behavior due to unexpected events, like the COVID-19 pandemic and new market regulations, 2) a quick equipment failure or 3) unforeseen weather conditions. 

\item \textbf{B. Gradual Drift:} Happens when the data patterns change progressively from the previous concept to the new one. This type shows a longer transition phase that involves a mixture of the old and new concepts. The new concept becomes more predominant over time.

\textbf{Examples:} 1) User preferences for a particular service or product can change slowly due to evolving trends, 2) the quality of machinery can degrade slowly over time, and 3) a patient's health may gradually change due to aging or the progression of a medical condition over the years. 

\item \textbf{C. Incremental Drift:} Occurs when the current concept replaces the past concept slowly, by involving intermediate concepts (may not be statistically significant). Some researchers consider incremental drift as a variant of gradual drift, however, what sets incremental drift apart from gradual drift is the absence of a distinct boundary that separates the old and new concepts. 

\textbf{Examples:} 1) The slow change in climate patterns (like in temperatures and rainfalls) progressively affect agricultural products, and 2) fraudsters in e-payment systems gradually adopt new tactics by shifting from simple techniques (like stolen credit cards) to modern ones (like virtual cards and bots). 

\item  \textbf{D. Recurrent Drift:} Happens when a concept that was encountered earlier reappears after some time has passed. While this shares similarities with gradual drift in that alteration of two concepts, the key distinction is that the old concept reappears after some time interval in the recurring drift.

\textbf{Examples:} 1) Seasonal changes in spending behavior, such as increased purchases during the holiday seasons, and 2) with the winter season comes health changes due to people becoming less active and having higher heart rates.
\end{itemize}

\section{Concept Drift Detection}
\subsection{Active vs. Passive Detection}
Drift detection methods can be mainly categorized into two folds \cite{Han2022, Suárez-Cetrulo2023}: active (with an explicit drift detection mechanism) vs. passive (continuous model adaptation). Active or informed detectors monitor data streams for drifts and activate some adaptation mechanisms when drifts have occurred, reducing FP and FN rates and saving memory and CPU resources \cite{Jain2022, Aguiar2024}.  Recent examples of active detectors include: 1) probabilistic methods that specifically identifies real drift \cite{Parasteh2024} and \cite{SirvanLast2025},  2) a sum-product NN-based method that detects both real and virtual drifts \cite{Parasteh2023b}, and 3) a cross-entropy-based method that identifies real drifts within noisy data streams \cite{Parasteh2023a} . On the other hand, passive methods continuously update their models with incoming samples without any drift detection because they consider that drifts may occur constantly or periodically. These methods can be handy in detecting gradual/incremental shifts, but they are time-consuming. An example is the incremental feature learning approach introduced in \cite{Sadreddin2022} that continuously adjusts a single NN model to new data chunks. The model is composed of several interconnected sub-NNs, with a new sub-NN optimally created for each incoming chunk. To prevent unbounded NN growth, only the sub-NNs most relevant to the new data distribution are retained and re-combined to build the optimal model.
\subsection{Learner-based vs. Distribution-based Detection}
Drift detectors can also be broadly classified into three main categories: learner-based (supervised), distribution-based (unsupervised) and hybrid. Each category has specific advantages and disadvantages \cite{lu2020a, gama2014, webb2017, zliobaite2010}, as explained below.
\begin{itemize}
\item \textbf{Learner-based:} These methods detect drifts by monitoring the performance of the underlying classifiers, such as the error rates. They are further split into three groups: Statistical Process Control (SPC), Windowing Techniques and Ensemble Learning. The pros of these methods are: 1) Directly linked to the predictive performance of the base models, which makes them intuitive and easy to interpret, and 2) Effective in scenarios where the drifts directly impact the model accuracy. Their cons are: 1) May not detect drifts that do not immediately affect model performance, such as changes in features, 2) Require data to be labeled, and 3) Can be sensitive to noise and random fluctuations in data.  
    
\item \textbf{Distribution-based:} These methods monitor changes in the data distribution itself over different timestamps to check whether the current and historical data windows come from the same distribution or not. They often use statistical tests (like Kullback-Leibler divergence and Kolmogorov-Smirnov Test) to determine if the data distribution has changed significantly. Their pros are: 1) Detect drifts that do not immediately affect model performance, 2) Often more robust to noise in data, and 3) Do not require data to be labeled. On the other hand, their cons are: 1) May require a large amount of data to detect drifts accurately, 2) More prone to false alarms, and 3) Can be computationally intensive due to continuous statistical testing.

\item \textbf{Hybrid:} Hybrid detectors combine elements of learner and distribution-based methods to leverage the strengths of both methods and provide more robust drift detection mechanisms. Their pros are: 1) Potentially more robust and versatile in different scenarios, and 2) Can dynamically switch between performance and distribution-based detection based on the context. Their cons are : 1) Can be more complex to implement and tune, and 2) May incur higher computational costs due to the dual nature of detection.
\end{itemize}
\subsection{Learner-based Detectors}
This category explicitly manages drift by incorporating mechanisms to forget old data and retrain or adjust the underlying models to the newly detected concept. To distinguish drift from noise, the new concept must remain stable for some time \cite{bayram2022}. The detectors are usually designed to be independent of specific ML algorithms, as they can adopt various types of learners. The general form of the learner-based detection methods is summarized in Algorithm 1 using the prequential training scheme (test and then train).  The algorithm is online, so that the drift is identified in real time.  The selected classifier is first pre-trained on an initial robust labeled dateset. The classifier then receives sequentially and continuously a stream of samples and predicts their labels. Next, it uses a statistical test method to detect a drift by comparing a drift metric to a predefined statistical threshold. For example, when the error rate (the discrepancy between predicted and true labels) increases significantly, this determines a drift.  Other examples would be comparing past accuracy or prediction probability (confidence score) with the accuracy or probability on the new data.  Some detectors use a warning threshold to signal an impending drift and monitor the samples more closely. 

Once a significant change is detected, the classifier retrains on recent data, typically stored in a sliding window or buffer. When the buffer reaches the limit, it forgets old data. Other methods select the most representative samples from the buffer or use all the samples but the new data have higher weights so that the detection model focuses less on old data \cite{iwashita2019}. 
\begin{algorithm}
    \caption{General Algorithm for Learner-based Drift Detection}\label{alg:cap}
    \begin{algorithmic}[1]
        \State \textbf{Inputs:} classier, intialDataset, dataStream (continous), statThreshold, RDSize (sliding window size)
        \State Pre-train classifier with intialDataset
        \State recentData $\gets$ empty /*window of recent data*/
            \For {each newSample in dataStream}
                \State Add newSample to recentData
                \If{(length of recentData $>$ RDSize) }
                  \State Remove oldest sample from recentData
                \EndIf 
                \State Predict sample's label with classifier
                \State Calculate driftMetric on prediction
                \If {(driftMetric $>$ statThreshold) /*significant difference*/}
                    \State Display "Drift Detected"
                    \State Train classifier on recentData with ground truth
                    \State recentData $\gets$  empty
                \EndIf 
            \EndFor
    \end{algorithmic}
\end{algorithm}
\subsection{A Summary} 
Table \ref{table:comparison-algorithm} compares numerous detectors (those selected for the experiments) based on their strategies: (1) Active (with an explicit drift detection mechanism) vs. Passive (continuous model adaptation),  (2) Online (processes each sample once and then discarded it) vs. Block-based (processes a data chunk at once), and (3) types of learner-based methods: SPC, Window techniques and Ensemble methods. 

\begin{table}[h!]
\centering
\small
\begin{tabular}{|l|l|l|l|l|}
 \hline
 \textbf{Method} &  \textbf{Year} & \textbf{Passive vs Active} & \textbf{Online vs Block} & \textbf{Category} \\
 \hline
 FTDD \cite{Cabral2018} & 2018 & Active & Online & SPC \\
 \hline
 RDDM \cite{Barros2017} & 2017 & Active & Online & SPC \\
 \hline
 FHDDM \cite{Pesaranghader2016} & 2016 & Active & Online & SPC \\
 \hline
 EWMA \cite{Ross2012} & 2012 & Active & Online & SPC \\
 \hline
 EDDM \cite{Baena-Garcıa2006} & 2006  & Active & Online & SPC \\
 \hline
 KSWIN \cite{Raab2020} & 2020 & Active & Online & Window \\
 \hline
 FPDD \cite{Cabral2018} & 2018 & Active & Online & Window \\
 \hline
 WSTD \cite{Barros2018} & 2018 & Active & Online & Window \\
 \hline
 MDDM \cite{Pesaranghader2018} & 2018 & Active & Online & Window \\
 \hline
 ADWIN \cite{Bifet2007} & 2007 & Active & Online & Window \\
 \hline
 D3 \cite{Gözüaçık2019} & 2015 & Active & Online & Window \\
 \hline
 ARF \cite{Gomes2017} & 2017 & Passive & Online & Ensemble \\
 \hline
 AUE \cite{Brzezinski2011} & 2011 & Passive & Block-based & Ensemble \\
 \hline
 DWM \cite{Kolter2007} & 2007 & Active & Online & Ensemble \\
 \hline
 AWE \cite{Wang2003} & 2003 & Active & Block-based & Ensemble \\
 \hline 
\end{tabular}

\caption{A Summary  of Learner-based Drift Detection Algorithms}
\label{table:comparison-algorithm}
\end{table}
 \section{SPC-based Detection}
SPC detectors treat a model's performance as a controlled process \cite{bayram2022, Nishida2007}. They employ statistical tests to detect deviations from expected behavior or exceedance of predefined control limits. By monitoring the evolution of a model's performance, SPCs assess the quality of the learning process \cite{bayram2022, Nishida2007}. When performance deteriorates or crosses a statistically significant threshold, it indicates concept drift. In such cases, SPCs may trigger classifier retraining or apply alternative strategies to address the drift. This approach enables real-time monitoring and facilitates prompt responses to changes in the underlying data. There are different SPC procedures, such as those defined in DDM, EDDM, RDDM, FHDDM, KAPPA, EWMA, STEPD, FTDD and ACDDM. We describe the most popular methods in the following sections. 

\subsection{EDDM}
Early Drift Detection Method (EDDM) \cite{Baena-Garcıa2006} is an improvement of the DDM \cite{gama2004}, designed to better detect gradual (moderate) changes.  Instead of considering the number of errors as in DDM, EDDM focuses on the distance between two successive errors.  At timestamp $t$, EDDM calculates \cite{Baena-Garcıa2006}: 
\begin{itemize}
\setlength\itemsep{0em}
    \item $p_{t}$, which is the average distance between two consecutive errors at time $t$
    \item $\sigma _{t}$,  which is the standard deviation of the distances at time $t$
    \item $p_{max}$ and $\sigma _{max}$, which track the maximum observed performance  and take the values of $p_{t}$ and $\sigma _{t}$ when $(p_{t} + 2\sigma _{t})$ reaches its historical maximum value
\end{itemize}

The study \cite{Baena-Garcıa2006} defines the warning and drift levels using $\alpha$ as the threshold for the warning zone and $\beta$ as the threshold for the drift zone:
\begin{itemize}
\setlength\itemsep{0em}
    \item Warning level: $(p_{t}  + 2\sigma _{t} ) / (p_{max} + 2\sigma _{max}) < \alpha $, indicating a possible change
    \item Drift level: $ (p_{t}  + 2\sigma{_t} ) / (p_{max} + 2\sigma_{max}) < \beta $, signaling a confirmed drift

\end{itemize}
The paper \cite{Baena-Garcıa2006} recommends setting $\alpha$ and $\beta$ to 0.95 and 0.90 respectively. 
When the warning level is detected, the samples are stored in preparation for drift localization. When the drift level is reached, the classifier along with $p_{max}$ and $\sigma_{max}$ are reset. In this case, the model can use the data collected in the warning zone together with recent post-drift data to train or update the model. When changes occur gradually at a moderate pace, EDDM is a better option than DDM. However, DDM is a better option when dealing with sudden changes. The limitation of EDDM can be pointed out as it is noise sensitive, which makes it more prone to false positives \cite{Nishida2007}. 

\subsection{FHDDM}
Fast Hoeffding Drift Detection Method (FHDDM) \cite{Pesaranghader2016} uses the Hoeffding inequality \cite{Hoeffding1963} to detect changes in data streams. It continuously monitors the prediction errors and uses the Hoeffding bound to determine whether the error rate has increased significantly or not. FHDDM computes the mean error in this window and compares it with the minimum error observed so far, and uses the following Hoeffding inequality: 
\[ p(|\bar{X} - \mu| \geq \epsilon) \leq 2e^{-2n\epsilon^2} \] 

where $\bar{X}$ is the average of the error rates of the current window, $\mu$ is the minimum error rate assuming no drift, $n$ is the size of the sliding window, and $\epsilon$ is the acceptable deviation from the minimum error rate. The method flags a drift when the deviation $|\bar{X} - \mu|$ is large enough, i.e. greater than or equal to the calculated threshold $\epsilon$, derived from the Hoeffding bound.

\subsection{RDDM}
Reactive Drift Detection Method (RDDM) enhances DDM by dynamically adjusting the detection threshold based on recent error rates \cite{Agrahari2022, Barros2017a}. RDDM utilizes the following strategy, based on the calculated error rate and standard deviation \cite{Barros2017a}: 
\begin{itemize}
\setlength\itemsep{0em}
    \item Warning level: $p_t + \sigma_t \geq p_{min} + (2 * \sigma_{min})$, for a possible drift;  $p_{min}$ is the historical minimum error rate and $\sigma_{min}$ is the standard deviation of $p_{min}$
    \item Drift level: $p_t + \sigma_t \geq p_{min} + (3 * \sigma_{min})$, for a detected drift
\end{itemize}

RDDM continuously recalculates warning and drift statistics, discarding outdated samples and focusing on recent ones. It identifies concepts that remain active for an extended period or with a long warning window. This adaptability allows to effectively identify both sudden and gradual drifts while minimizing false alarms \cite{Barros2018, Suárez-Cetrulo2023}. Compared to DDM, RDDM achieves higher accuracy in most scenarios, particularly for gradual drifts, by detecting more drifts and identifying them earlier. However, its increased complexity and parameter sensitivity may require careful calibration for optimal performance in specific applications \cite{bayram2022}.


\subsection{EWMA}
The exponentially weighted moving average (EWMA) was proposed in \cite{Ross2012}. EWMA for drift detection is a statistical process that identifies drift by monitoring the error rate of a streaming classifier \cite{barros2019, Cabral2018}.The standard deviation of the EWMA estimator {$z_t$} is \cite{lu2020a}: \[ \sigma_{z_t}^2 = \sqrt{p_{0}(1 - p_{0}) \frac{\lambda}{2 - \lambda} (1 - (1 - \lambda)^{2t}) } \]

Here, $ z_t = (1 - \lambda)z_{t-1} + \lambda x_t $, $p_{0}$ is the baseline error rate under no-drift conditions, and $\lambda$ is a smoothing parameter.\\

According to EWMA, drift is detected if \cite{bayram2022}: $z_t > \mu_{0} + L\sigma_{z_t}$. 
Here, $L$ is the control limit, which states the distance $z_t$ must deviate from $\mu _0$ before a change is flagged. Setting the appropriate smoothing factor ($\lambda$) is crucial for achieving optimal performance. $\lambda = 0.2$ has been suggested  by \cite{lu2020a}. A high $\lambda$ value increases sensitivity to recent changes but may make the method susceptible to noise. A low $\lambda$ value can lead to delayed drift detection\cite{Ross2012}. The main advantage of EWMA is its ability to detect gradual and abrupt drifts \cite{Agrahari2022}. It is also computationally efficient, with an overhead of $O(1)$ \cite{Ross2012}.
\subsection{FTDD}\label{sec:ftdd_spc}
The Fisher Test Drift Detector (FTDD) is one of three drift detection algorithms proposed by \cite{Cabral2018}, based on an efficient implementation of Fisher's Exact Test \cite{Fisher1922}. FTDD is based on the Sequential Testing with Estimation of Proportion Difference (STEPD) algorithm \cite{Nishida2007}, but addresses its limitations when dealing with imbalanced data \cite{wares2019} Unlike STEPD, which uses a test of equal proportions, FTDD employs Fisher's Exact test in all comparisons to detect concept drift \cite{Barros2018, barros2019, bayram2022}. The $p_{value}$ calculation is a key feature of FTDD, contributing to its effectiveness in accurately detecting drifts \cite{Cabral2018}: 

\[p\_{value} = \frac{\frac{|w_r + w_p|}{|w_r|}}{|w_p|} \times \frac{\frac{|c_r + c_p|}{|c_r|}}{|c_p|} \]
\[ p\_{value} = p\_{value} \times constF \times 2 \]

Where $w_r$ is the number of errors in the recent window, $w_p$ is the proportionally calculated number of errors in past window, $c_r$ is the number of correct predictions in the recent window, $c_p$ is the proportionally calculated number of correct predictions in past window. $constF$ is a constant factor derived from the factorial calculations for the window size ($w$). It helps to optimize the $p\_{value}$ computation by pre-calculating and storing a common factor. Here, the probability of observing the specific distribution of errors and correct predictions in the two windows (recent and past) is calculated given the assumption that no drift has occurred \cite{Cabral2018}. The warning is triggered when the $p\_{value}$ falls below the threshold $\alpha_w = 0.005$ and a drift is confirmed when the $p_{value}$ is typically less than $\alpha_d = 0.003$.

\section{Window-Based Detection}
Windowing detectors handle drifts by monitoring the model performance or other statistical measures over two data windows (of a fixed or dynamic size) and checking for significant discrepancies between them \cite{bayram2022, wares2019}.  They treat the classifier's prediction results as the data stream. They assess the chosen metric on each window and flag drift when the difference exceeds a predefined threshold or a threshold derived from a statistical test \cite{bayram2022, wares2019}. The dynamic windowing technique adjusts the size based on the drift's length, retaining samples until the drift occurs. It allows for flexibility without the need to set a predefined window size, which is a difficult task. The decision for window size is crucial because large windows can increase accuracy but may miss the detection of rapid drifts, and small windows are more effective in detecting sudden drifts but might not capture gradual changes. There are different windowing strategies, as described in the following selected methods: ADWIN, KSWIN, MDDM, FPDD, WSTD and D3. 
\subsection{ADWIN}
ADaptive WINdowing (ADWIN) \cite{Bifet2007} maintains a sliding window $w$ containing the most recent errors and tests, after every new instance, whether the average error in an older window $w_p$ differs significantly from that in a newer window $w_r$. ADWIN detects drift by comparing the mean error rates ($\mu_{w_p}$ for $w_p$, $\mu_{w_r}$ for $w_r$) of the two sub-windows. A drift is signaled when the difference in error rates exceeds a threshold derived from the Hoeffding bound \cite{wares2019, bayram2022}: $|\mu_{w_p} - \mu_{w_r}| > \epsilon$, where $\epsilon$ is the threshold calculated based on the Hoeffding bound \cite{Bifet2007}:

\[ \epsilon = \sqrt{\frac{1}{2m}ln(\frac{4|W|}{\delta})} \]

Where $m$ is the harmonic mean of the lengths of sub-windows $w_{r}$ and $w_{p}$, $|w|$ is the length of the entire window $w$, $\delta$ is a predefined confidence parameter, and $ln()$ is the natural logarithm. ADWIN has been shown to be effective in detecting both abrupt and gradual concept drifts \cite{Mehdi2024}. 

\subsection{KSWIN}
Kolmogorov–Smirnov Windowing (KSWIN) \cite{Raab2020} maintains two sliding windows of error observations, an older window, $w_p$, and a recent window, $w_r$ \cite{Parasteh2023a, Nunes2024}.  After each new prediction, the sup-norm KS (Kolmogorov–Smirnov) distance, $dist(w_r, w_c)$ is computed for sample $x$, where sup-norm distance is the maximum point-wise gap between two functions, making it highly sensitive to even a single large deviation and thus effective for detecting distribution changes \cite{Raab2020}: 

\[ dist(w_p, w_r) = sup|f_{w_p}(x) - f_{w_r}(x)| \]

Here, $f(x) = \frac{1}{n} \sum_{i=1}^{n} I_{(-\infty, x)} (x_i)$ and $I_{(-\infty, x)} (x_i)$ is an indicator function. \\

A drift is signaled by KSWIN when the calculated distance exceeds a threshold defined by the formula below \cite{Parasteh2023a, Raab2020}:

\[ dist(w_p, w_r) > \sqrt{-\frac{ln(\alpha)}{n}} \]

Where $\alpha$ is the significance level representing the probability of incorrectly rejecting the null hypothesis. Common values for $\alpha$ are 0.05, 0.01, or even smaller depending on the desired level of confidence, and $n$ is the size of the two sub-windows \cite{Raab2020}.

\subsection{MDDM}
McDiarmid Drift Detection Method (MDDM) \cite{Pesaranghader2018} uses the McDiarmid’s inequality introduced in \cite{McDiarmid1989}. MDDM operates under the assumption that in a streaming environment, recent samples are more relevant than older ones. To reflect this, MDDM uses a sliding window with a weighting scheme that prioritizes the most recent samples \cite{Pesaranghader2018}. MDDM slides a window of size $n$ over the prediction results \cite{Gulcan2022}. This window stores binary values, where $1$ represents an incorrect prediction and $0$ represents a correct prediction \cite{bayram2022}. Each element in the window is associated with a weight, where $w_i < w_{i+1}$. While inputs are processed, the current weighted average of the elements of the window is calculated, i.e. $\mu^t$, as well as the maximum weighted mean observed so far, i.e. $\mu^m$. When $ \mu^m < \mu^t $ then $ \mu^m = \mu^t $ \cite{wares2019, Pesaranghader2018}.  Hence, with MDDM, concept drif is declared when \cite{Pesaranghader2018}: 

\[ (\mu^m - \mu^t) \geq \epsilon \]

where $\epsilon$ is the threshold calculated with the McDiarmid's inequality: 
\[ \epsilon = \sqrt{\frac{\sum_{i=1}^{n} v_i^2}{2} * ln(\frac{1}{\delta})} \]

Here, $\delta$ is the confidence level, and $ v_i$ is the weight of the i-th element, calculated as: $ v_i = \frac{w_i}{\sum_{i=1}^{n} w_i} $ \cite{Pesaranghader2018}.
\subsection{FPDD}
The Fisher Proportions Drift Detector (FPDD)\cite{Cabral2018} is designed to enhance the performance of the Statistical Test of Equal Proportions(STEPD), particularly when dealing with small sample sizes \cite{bayram2022}. It leverages Fisher's Exact Test to provide a more robust statistical analysis in these scenarios \cite{barros2019}. This leads to more accurate drift detection when drifts occur rapidly. Similar to STEPD, FPDD compares two windows, the recent window ($w_r$) and past window ($w_p$), to analyze the data stream. 

The core distinction of FPDD lies in its conditional application of Fisher's Exact Test. Suppose the number of errors or correct predictions in either of the two windows is smaller than five. In this case, FPDD uses Fisher's Exact Test to assess the statistical significance of the difference in error rates between the two windows. In cases where the sample sizes are sufficient(5 or more), FPDD reverts to the standard test of equal proportions used by STEPD\cite{wares2019}. Based on the chosen statistical test (Fisher's Exact Test or the test of equal proportions), FPDD calculates the $p\_{value}$ (as computed in section \ref{sec:ftdd_spc}), and uses it as follows \cite{Cabral2018}: 
\begin{itemize}
    \item Drift detected: if $ p\_{value} \leq \alpha_d$
    \item Warning signal: if $ \alpha_d < p\_{value} \leq \alpha_w $ 
\end{itemize}

FPDD uses two similar threshold parameters like STEPD, the significance levels for the detection of drifts ( $\alpha_d = 0.003$ ) and warning ( $\alpha_w = 0.05$ ).

\subsection{WSTD}
Wilcoxon Rank Sum Test Drift Detector (WSTD) \cite{Barros2017}, which closely resembles STEPD, detects drifts based on an efficient implementation of the Wilcoxon rank sum statistical test \cite{Wilcoxon1945}, whereas STEPD uses the test of equal proportions \cite{Barros2018, Han2022}. WSTD requires setting a significance level ($\alpha$) and applies the normal distribution to evaluate the null hypothesis. Given two sample set, $n_1$ and $n_2$, they are combined  in the ascending order as below \cite{Barros2017}: 
\begin{itemize}
    \setlength\itemsep{-0.2em}
    \item Test statistic is $z = \frac{(R - \mu_R)}{\sigma_R}$
    \item Population mean is $\mu_R = n_1 \times \frac{(n_1 + n_2 +1)}{2}$
    \item Standard deviation is $\sigma_R = \sqrt{n_1 \times n_2 \times \frac{(n_1 + n_2 + 1)}{12}}$
\end{itemize}

where $R$ is the smallest sum of the ranks of both sample sets, $n_1$ is the size of the smallest sample set, and $n_2$ is the size of the largest sample set. The $z$ value is used to reject the null hypothesis, and the $p\_value$ (or obtained probability) is also required to find the $z$ value \cite{Agrahari2022, Barros2017}. 

Similar to STEPD, WSTD monitors the base learner’s predictions using two windows: a recent window and an older one. It relies on a statistical test to issue warnings or confirm drifts. WSTD incorporates three key parameters with default values: the recent window size ($w = 30$), the significance level for drift detection ($\alpha_d = 0.003$), and the significance level for warnings ($\alpha_w = 0.05$) \cite{Barros2017}.
\subsection{D3}
Discriminative Drift Detector (D3) \cite{Gözüaçık2019} detects changes by comparing recent and historical samples. D3 operates on the principle that a change in the underlying data distribution $P(X)$ will manifest as a difference between recent and historical data. By training a classifier to distinguish between these sets, D3 can detect drifts when the classifier performs well, which suggests that the data distributions have diverged. D3 employs a sliding window, denoted as $W$, to store the most recent samples. The size of this window is determined by $w(1 + \rho)$, where \cite{Gözüaçık2019}:

\begin{itemize}
\setlength\itemsep{-0.2em}
    \item $w$ is the number of "old" samples, which correspond to the historical data
    \item $\rho$ is the fraction of "new" samples, relative to the size of the old data. This essentially controls the size of the new dataset within the window
    
\end{itemize}

A drift is signaled when the classifier AUC score (trained to distinguish between the old and new sample) is greater than or equal to the threshold, $\theta$, which is typically set between 0.5 and 1.0 \cite{Schwengber2022, Gözüaçık2019}. The performance of D3 can vary significantly based on the specific classifier and dataset being used \cite{Hinder2023}. Additionally, the method's capability is constrained to identifying only linear drift patterns within the feature space \cite{Gözüaçık2020}. 

\section{Ensemble-based Detection}
Ensemble-based detectors are a robust approach to handling drifts because combining the outputs of several learners is more effective to changing data patterns. Many such detectors adopt the Weighted Majority Algorithm (WMA). The ensemble remains up to date with new concepts as follows \cite{bayram2022, Mehdi2024}: (1) a new classifier trained on the current data is added to the ensemble,  (2) the learners are then weighted according to their  performance with  the current data, and (3) the least-weighted learners, deemed less effective, are removed from the ensemble.  In addition to adding/removing learners, the ensemble may use incremental learning to adapt the learners. The ensemble approach is efficient in identifying gradual drifts while maintaining high levels of accuracy \cite{bayram2022, Mehdi2024}, but comes with high computational and memory costs. There are numerous ensemble-based detectors, which can be categorized as follows: (1) methods that implement their own detection mechanisms, such as AWE, AUE, ACE and DWM, and (2) methods that utilize an existing detector, such as ADWIN, which is used in algorithms like ARF. We describe these methods in the following sections.

\subsection{AWE}
Accuracy Weighted Ensemble (AWE) \cite{Wang2003} selects the most efficient classifiers using the mean squared error (MSE), which is calculated based on the probabilities of class assignments. AWE identifies the top \textit{n} classifiers based on their performance on the most recent data chunk \cite{Wang2003}. It employs MSE for assigning weights: classifiers that exhibit an error rate meeting or exceeding a certain threshold are excluded. This method ensures that only the classifiers  most aligned with the current data patterns are retained, thereby enhancing the ensemble's overall accuracy. The weight ($w_i$) assigned to a classifier $c_i$ is calculated using the following formulas \cite{bayram2022, wares2019, Kolter2007}: 

\[w_i = MSE_r - MSE_i\]

Here $MSE_r$ is the reference MSE calculated based on the class distribution of the current data chunk $S_n$ \cite{Wang2003}:  

\[ MSE_r = \sum_c p(c)(1 - p(c))^2\]

where $p(c)$ is the probability of class $c$, which is estimated from the distribution of classes in $S_n$. $MSE_i$ represents the error for the $i^{th}$ classifier, calculated over $S_n$ as follows \cite{Wang2003}: 
\[ MSE_i = \frac{1}{S_n} \sum_{(x,c) \epsilon S_n}(1 - p_c^i(x))^2  \]

where  $p_c^i(x)$ is the probability that classifier $i$ assigns sample $x$ to class $c$. 
\subsection{AUE}
Accuracy Updated Ensemble (AUE) \cite{Brzezinski2011}, an advancement over AWE, updates only the top $k$ weighted classifiers that meet a certain accuracy threshold on the most recent data block, rather than updating all classifiers. Additionally, AUE utilizes a more straightforward weighting function compared to AWE \cite{wares2019}. AUE incorporates online classifiers, typically Very Fast Decision Trees (VFDT) or Hoeffding Trees, by updating them individually in addition to weight modifications. This adaptability ensures that in the absence of drift across chunks, the classifiers enhance their performance as though they were trained on a singular, extensive dataset \cite{wares2019}. Consequently, this flexibility permits a reduction in chunk size without compromising the ensemble's accuracy. The weighting function employed in AUE is simplified as below\cite{Brzezinski2011}:

\[ w_i = \frac{1}{MSE_i + \epsilon} \]

where $MSE_i$ is computed in the same manner as in AWE, and $\epsilon$ is a small constant added to facilitate weighting calculations even when $MSE_i$ equals zero \cite{Brzezinski2011}. The ensemble size in AUE is typically fixed to a predetermined number, $k$, of classifiers \cite{wares2019}. Comparative experimental studies have demonstrated that AUE outperforms AWE in various datasets, except for one instance where both achieved comparable accuracy \cite{Brzezinski2011, wares2019}. 


\subsection{ARF}
The Adaptive Random Forest (ARF) algorithm \cite{Gomes2017} handles evolving data streams by incorporating several key strategies. ARF extends the traditional random forest algorithm to adapt to changes in the data distribution by replacing under-performing trees with new ones \cite{Jiao2022, lu2020a}. 
ARF uses Hoeffding Trees (HT) as its base learners, which are a type of very fast decision tree. These trees are incrementally updated with each new sample, making them suitable for streaming data \cite{Gomes2017, Brzezinski2011}.  ARF employs a resampling method based on online bagging \cite{Suárez-Cetrulo2023}. Instead of growing each tree sequentially on different subsets of data, as in traditional random forests, ARF simulates sampling with reposition using a Poisson distribution, specifically Poisson ($\lambda = 6$) \cite{Gomes2017}. ARF handles concept drift through a combination of online bagging for resampling, drift detectors for each tree, background training of new trees, and weighted majority voting. These mechanisms allow the model to maintain high accuracy when dealing with non-stationary data streams\cite{Han2022}.

\subsection{DWM}
The Dynamic Weighted Majority (DWM)\cite{Kolter2007} extends the weighted majority algorithm \cite{Littlestone1994} by dynamically adjusting the weights of its base classifiers and adding/removing classifiers  based on their performance \cite{Krawczyk2017}. DWM is a passive adaptation method that does not use an explicit drift detection mechanism, but rather continuously adapts its base learner \cite{khamassi2018}. This differentiates it from methods that explicitly monitor error rates or other statistics to detect drift.

With DWM, weights are adjusted after each new classified sample. This online approach allows DWM to adapt to gradual changes more easily \cite{bayram2022, Han2022}:
\begin{itemize}
    \item If a base classifier predicts correctly, its weight is multiplicatively increased by a factor $\beta > 1$  i.e.,  $w_i(t+1) = \beta * w_i(t)$ \cite{wares2019}
    
    \item If a base classifier predicts incorrectly, its weight is decreased multiplicatively by a factor $\alpha < 1$  i.e., $w_i(t+1) = \alpha * w_i(t)$ \cite{wares2019}
\end{itemize}

where $w_i(t)$ is the weight of classifier $i$ at time $t$, and $w_i(t+1)$ is the updated weight at time $t+1$. Typical values for the parameters can be $\beta = 1.1$ and $\alpha = 0.9$.


\section{Experiments Setup}
We utilize both artificial and real-world datasets from the Harvard Dataverse \cite{harvarddataverse}, under abrupt and gradual scenarios, to assess the performance of drift detectors of different categories. These datasets present distinct data characteristics, providing an extensive testbed. When the drift is identified, the base learners, Naive Bayes and Hoeffding Tree, are retrained with the most recent samples to ensure that predictions remain accurate.  We assess and compare these detector's performance using the AUC metric.  AUC is useful for assessing performance degradation caused by drifts. For each detector's hyperparameters, we use their default values mentioned in earlier sections. In all the experiments, we set the window size to 50 and the ensemble size to 15. In the empirical tables, the best results are shown in red and the worst ones in blue.

\subsection{Implementation Frameworks}
Table \ref{table:Implementations-drift-detectors} provides the implementation frameworks of the 15 selected drift detectors. These frameworks support a broad collection of ML algorithms, such as classification, regression, clustering and drift detection. Here, NA indicates that there was no publicly available implementations found online. Therefore, we implemented the Python code with the assistance from Github Copilot. 

\begin{table}[h!]
    \centering
    \caption{Drift Detectors and their Implementation Frameworks}
    \label{table:Implementations-drift-detectors}
    \begin{small}
    \begin{tabular}{|l|l| }
        \hline
         & \textbf{URL} \\
         \hline
         \textbf{FTDD} \cite{Cabral2018} &  \href{https://github.com/brentp/fishers_exact_test.git}{Git (Fisher's Exact Test}\\
         \hline 
         \textbf{RDDM} \cite{Barros2017} & \href{https://github.com/alipsgh/tornado/blob/master/drift_detection/rddm.py}{Tornado}\\ 
         \hline
         \textbf{FHDDM} \cite{Pesaranghader2016} & \href{https://riverml.xyz/latest/api/drift/binary/FHDDM/}{River} \\
         \hline
         \textbf{EWMA} \cite{Ross2012} & \href{https://github.com/alipsgh/tornado/blob/master/drift_detection/ewma.py}{Tornado} \\
         \hline
         \textbf{EDDM} \cite{Baena-Garcıa2006} & \href{https://riverml.xyz/0.13.0/api/drift/EDDM/}{River} \\
         \hline
         \textbf{KSWIN} \cite{Raab2020} & \href{https://riverml.xyz/latest/api/drift/KSWIN/}{River} \\
         \hline
         \textbf{FPDD} \cite{Cabral2018} & \href{https://github.com/brentp/fishers_exact_test.git}{Git (Fisher's Exact Test)}\\
         \hline
         \textbf{WSTD} \cite{Barros2018} &  NA\\
         \hline
         \textbf{MDDM} \cite{Pesaranghader2018} & \href{https://github.com/alipsgh/tornado/blob/master/drift_detection/mddm_a.py}{Tornado} \\
         \hline
         \textbf{ADWIN} \cite{Bifet2007} & \href{https://riverml.xyz/latest/api/drift/ADWIN/}{River} \\
         \hline
         \textbf{ARF} \cite{Gomes2017} & \href{https://riverml.xyz/0.11.1/api/ensemble/AdaptiveRandomForestClassifier/} {River} \\
         \hline
         \textbf{D3} \cite{Gözüaçık2019} & \href{https://github.com/ogozuacik/d3-discriminative-drift-detector-concept-drift/blob/master/D3.py}{Git} \\
         \hline
         \textbf{AUE} \cite{Brzezinski2011} & NA \\
         \hline
         \textbf{DWM} \cite{Kolter2007} & \href{https://scikit-multiflow.readthedocs.io/en/stable/api/generated/skmultiflow.meta.DynamicWeightedMajorityClassifier.html}{Scikit-multiflow} \\
         \hline
         \textbf{AWE} \cite{Wang2003} & \href{https://scikit- multiflow.readthedocs.io/en/stable/api/generated/skmultiflow.meta.AccuracyWeightedEnsembleClassifier.html}{Scikit-multiflow} \\
         \hline
    \end{tabular}
    \end{small}
\end{table}

\subsection{Streaming Datasets}
Our empirical study utilizes synthetic and real datasets, summarized in Table \ref{table:datasets-configuration}, to evaluate the performance of various drift detectors. The six synthetic datasets (balanced binary class label and without noise) have been produced using three main generators, Random Tree Stream (RT), SINE and MIXED, by incorporating abrupt and gradual drifts \cite{harvarddataverse}:  
\begin{itemize}
\item RT-abrupt and RT-gradual, with the arrangement of 8873985678962563
\item Sine-abrupt and Sine-gradual, with the arrangement of 0123
\item Mixed-abrupt and Mixed-gradual, with the arrangement of 0101
\end{itemize}

Each dataset possess four distinct concepts and three drift locations \cite{harvarddataverse}. The sudden datasets experience abrupt shifts at locations of 10,000, 20,000 and 30,000. The gradual datasets undergo gradual transitions at locations of 9,500, 20,000 and 30,500, with each transition having a drift width of 1,000 samples. 
\medskip

On the other hand, we employ two real datasets (balanced binary class label), as presented in Table \ref{table:datasets-configuration}. The Electricity dataset \textbf{ELE}C2, collected from the Australian New South Wales electricity market, predicts whether the prices (set every five minutes) of electricity will increase or decrease \cite{realharvarddataverse}. The Intrusion Detection Evaluation dataset \textbf{CIC}-IDS2017 contains network traffic collected for five days, from Monday, July 3, 2017, at 9am to Friday, July 7, 2017, at 5pm \cite{Sharafaldin2018}.

\renewcommand{\arraystretch}{1.2}
\begin{table}[h!]
    \centering
    \begin{small}
    \begin{tabular}{|l|l|l|l|}
        \hline
         \textbf{Dataset} & \textbf{Size} & \textbf{\#Features-Modality} & \textbf{\#Drifts}   \\
         \hline
         $RT_a$ & 40,000 & 2-numerical &  3 Abrupt\\
         \hline
         {$RT_g$} & 40,000 & 2-numerical &  3 Gradual   \\
         \hline
         {$Sine_a$} & 40,000 & 2-numerical& 3 Abrupt   \\
         \hline
         {$Sine_g$} & 40,000 & 2-numerical & 3 Gradual  \\
         \hline
         {$Mixed_a$} & 40,000 & 4-numerical&  3 Abrupt  \\
         \hline
         {$Mixed_g$} & 40,000 & 4-numerical &  3 Gradual   \\
         \hline
         {ELE} & 45,312 & 8-mixed (6 numerical, 2 categorical)   &  Unknown\\
         \hline
         {CIC} & 28,303 & 19-mixed (16 numerical, 3 categorical)    & Unknown\\
         \hline
    \end{tabular}
    \end{small}
    \caption{Configurations of Synthetic and Real-world Datasets}
    \label{table:datasets-configuration}
\end{table}

\section{Comparison of SPC-based Detectors}
For SPC detectors (cf. Table \ref{table:Comparisons-50-window-SPC}), on abrupt-drift datasets, NB shows modest performance, with FTDD achieves the highest AUC average, while the other methods all yield a lower average. With HT, performance is considerably higher across all detectors, confirming the advantage of the more expressive base learner. FTDD performs best, followed by EWMA, while RDDM shows the lowest performance.  On gradual-drift datasets, NB performance remains uniform across all SPC methods, where HT yields moderately higher values and EWMA and EDDM tied at the top.  On real-world data streams, NB reveals a notable divergence among detectors: FTDD records the lowest average, whereas the remaining methods achieve a substantially higher average. For HT, the results follow a similar pattern: FTTD yields the lowest average AUC, whereas the other methods achieve comparable performance.

When considering all dataset types together, overall HT surpasses NB.  The results indicate that while FTDD holds an advantage on abrupt-drift data, EWMA and EDDM deliver the most reliable performance across all dataset types.

\begin{table}[h!]
    \small
    \centering
    \setlength{\tabcolsep}{2pt} 
    \renewcommand{\arraystretch}{1.2} 
\begin{tabular}{ |p{1.25cm}||p{1.1cm}|p{1.1cm}|p{1.3cm}|p{1.1cm}|p{1.1cm}|p{1.1cm}|p{1.1cm}|p{1.3cm}|p{1.1cm}|p{1.1cm}|  }
 \hline
  \multicolumn{1}{|c|}{} &
  \multicolumn{5}{|c|}{NB} &
  \multicolumn{5}{|c|}{HT} \\
 \hline
 \textbf{} & FTDD & RDDM & FHDDM & EWMA & EDDM & FTDD & RDDM & FHDDM & EWMA & EDDM \\
 \hline
 \hline
 \textbf{$RT_a$} & \cellcolor[HTML]{7EA6E0} 0.57 & 0.62 & \cellcolor[HTML]{FF9999} 0.63 & \cellcolor[HTML]{FF9999} 0.63 & \cellcolor[HTML]{FF9999} 0.63 & \cellcolor[HTML]{7EA6E0} 0.60 & 0.70 & 0.74 & 0.75 & \cellcolor[HTML]{FF9999} 0.76 \\
 \hline
 \textbf{$Sine_a$} & \cellcolor[HTML]{FF9999} 0.61 & 0.52 & \cellcolor[HTML]{7EA6E0} 0.51 & \cellcolor[HTML]{7EA6E0} 0.51 & \cellcolor[HTML]{7EA6E0} 0.51 & \cellcolor[HTML]{FF9999} 0.78 & \cellcolor[HTML]{7EA6E0} 0.50 & 0.56 & 0.56 & 0.56 \\
 \hline
 \textbf{$Mixed_a$} & \cellcolor[HTML]{FF9999} 0.55 & \cellcolor[HTML]{7EA6E0} 0.52 & \cellcolor[HTML]{7EA6E0} 0.52 & \cellcolor[HTML]{7EA6E0} 0.52 & \cellcolor[HTML]{7EA6E0} 0.52 & \cellcolor[HTML]{FF9999} 0.69 & 0.55 & 0.56 & 0.57 & \cellcolor[HTML]{7EA6E0} 0.54 \\
 \hline
\textbf{$Aver_a$} & \textbf{0.58} & \textbf{0.55} & \textbf{0.55} & \textbf{0.55} & \textbf{0.55} & \textbf{0.69} & \textbf{0.58} & \textbf{0.62} & \textbf{0.63} & \textbf{0.62} \\

 \hline
 \hline
 
  \textbf{$RT_g$} & \cellcolor[HTML]{FF9999} 0.63 & \cellcolor[HTML]{FF9999} 0.63 & \cellcolor[HTML]{FF9999} 0.63 & \cellcolor[HTML]{FF9999} 0.63 & \cellcolor[HTML]{FF9999} 0.63 & \cellcolor[HTML]{7EA6E0} 0.76 & 0.77 & 0.77 & \cellcolor[HTML]{FF9999} 0.78 & \cellcolor[HTML]{FF9999} 0.78 \\
 \hline
 \textbf{$Sine_g$}  & \cellcolor[HTML]{7EA6E0} 0.50 & \cellcolor[HTML]{7EA6E0} 0.50 & \cellcolor[HTML]{FF9999}  0.51 & \cellcolor[HTML]{FF9999} 0.51 & \cellcolor[HTML]{FF9999} 0.51 & \cellcolor[HTML]{7EA6E0} 0.49 & 0.50 & \cellcolor[HTML]{7EA6E0} 0.49 & \cellcolor[HTML]{FF9999} 0.52 & \cellcolor[HTML]{FF9999} 0.52 \\
 \hline
 \textbf{$Mixed_g$} & \cellcolor[HTML]{FF9999} 0.51 & \cellcolor[HTML]{FF9999} 0.51 & \cellcolor[HTML]{FF9999} 0.51 & \cellcolor[HTML]{FF9999} 0.51 & \cellcolor[HTML]{FF9999} 0.51 & \cellcolor[HTML]{FF9999} 0.50 & \cellcolor[HTML]{7EA6E0} 0.47 & 0.49 & 0.49 & 0.49 \\
 \hline
\textbf{$Aver_g$} & \textbf{0.55} & \textbf{0.55} & \textbf{0.55} & \textbf{0.55} & \textbf{0.55} & \textbf{0.58} & \textbf{0.58} & \textbf{0.58} & \textbf{0.60} & \textbf{0.60} \\

 \hline
 \hline
 
 \textit{ELE}  & \cellcolor[HTML]{7EA6E0} 0.70 & \cellcolor[HTML]{FF9999} 0.75 & \cellcolor[HTML]{FF9999} 0.75 & \cellcolor[HTML]{FF9999} 0.75 & \cellcolor[HTML]{FF9999} 0.75 & \cellcolor[HTML]{7EA6E0} 0.71 & \cellcolor[HTML]{FF9999} 0.75 & \cellcolor[HTML]{FF9999} 0.75 & \cellcolor[HTML]{FF9999} 0.75 & \cellcolor[HTML]{FF9999} 0.75 \\
 \hline
\textit{CIC}  & \cellcolor[HTML]{7EA6E0} 0.38 & \cellcolor[HTML]{FF9999} 0.80 & \cellcolor[HTML]{FF9999} 0.80 & \cellcolor[HTML]{FF9999} 0.80 & \cellcolor[HTML]{FF9999} 0.80 & \cellcolor[HTML]{7EA6E0} 0.71 & \cellcolor[HTML]{FF9999} 0.78 & \cellcolor[HTML]{FF9999} 0.78 & \cellcolor[HTML]{FF9999} 0.78 & 0.77 \\
 \hline
\textbf{$Aver_{r}$} & \textbf{0.54} & \textbf{0.78} & \textbf{0.78} & \textbf{0.78} & \textbf{0.78} & \textbf{0.71} & \textbf{0.77} & \textbf{0.77} & \textbf{0.77} & \textbf{0.76} \\
 \hline
\end{tabular}
    \caption{Comparisons of SPC-based Methods Combined With Two Base Learners}
    \label{table:Comparisons-50-window-SPC}
\end{table}
\section{Comparison of Window-based Detectors}

For window-based detectors (cf. Table \ref{table:Comparisons-50-window-Window}), on abrupt-drift datasets, NB shows uniform and modest performance, with all six methods yielding an identical average AUC, offering no differentiation among detectors. With HT, performance is consistently higher and slight differences emerge: KSWIN, WSTD and D3 each reach the same top average value, while FPDD, MDDM and ADWIN obtain a marginally lower average.

On gradual-drift datasets, NB again remains stable and uniform across all methods. HT yields moderately higher values, where WSTD, MDDM, ADWIN and D3 all reach the same top average, while KSWIN and FPDD each fall slightly behind.

On real-world data streams, NB reveals a notable divergence among detectors: KSWIN records the lowest average, driven by a sharp drop on the CIC dataset, whereas the remaining methods achieve a substantially higher average. For HT, the results follow a similar pattern: KSWIN yields the lowest average, while the other methods achieve comparable or slightly higher performance.

When considering all dataset types together, HT surpasses NB across all window-based methods, though the margin remains modest. Overall, window-based methods yield similar aggregate performance levels to SPC detectors, and no single window-based detector emerges as clearly dominant within either base learner when all dataset types are combined. The results indicate that while KSWIN, WSTD and D3 share an advantage on abrupt-drift data, WSTD and D3 deliver the most reliable performance across all dataset types.

\begin{table}[h!]
    \small
    \centering
    \setlength{\tabcolsep}{2pt} 
    \renewcommand{\arraystretch}{1.2} 
\begin{tabular}{ |p{1.25cm}||p{1.2cm}|p{1.0cm}|p{1cm}|p{1.2cm}|p{1.2cm}|p{0.7cm}|p{1.2cm}|p{1cm}|p{1cm}|p{1.2cm}|p{1.2cm}|p{0.7cm}|  }
 \hline
  \multicolumn{1}{|c|}{} &
  \multicolumn{6}{|c|}{NB} &
  \multicolumn{6}{|c|}{HT} \\
 \hline
 \textbf{} & KSWIN & FPDD & WSTD & MDDM & ADWIN & D3 & KSWIN & FPDD & WSTD & MDDM & ADWIN & D3 \\
 \hline
 \hline
 \textbf{$RT_a$} & \cellcolor[HTML]{7EA6E0} 0.63 & \cellcolor[HTML]{7EA6E0} 0.63 & \cellcolor[HTML]{FF9999} 0.64 & \cellcolor[HTML]{7EA6E0} 0.63 & \cellcolor[HTML]{7EA6E0} 0.63  & \cellcolor[HTML]{7EA6E0} 0.63 & \cellcolor[HTML]{7EA6E0} 0.77 & \cellcolor[HTML]{7EA6E0} 0.77 & \cellcolor[HTML]{FF9999} 0.78 & \cellcolor[HTML]{FF9999} 0.78 & \cellcolor[HTML]{FF9999} 0.78 & \cellcolor[HTML]{FF9999} 0.78  \\
 \hline
 \textbf{$Sine_a$} & \cellcolor[HTML]{FF9999} 0.51 & \cellcolor[HTML]{7EA6E0} 0.50 & \cellcolor[HTML]{7EA6E0} 0.50 & \cellcolor[HTML]{7EA6E0} 0.50 & \cellcolor[HTML]{7EA6E0} 0.50 & \cellcolor[HTML]{7EA6E0} 0.50 & \cellcolor[HTML]{FF9999} 0.54 & 0.53 & \cellcolor[HTML]{FF9999} 0.54 & \cellcolor[HTML]{7EA6E0} 0.51 & 0.52 & 0.53 \\
 \hline
 \textbf{$Mixed_a$} & \cellcolor[HTML]{FF9999} 0.52 & \cellcolor[HTML]{FF9999} 0.52 & \cellcolor[HTML]{FF9999} 0.52 & \cellcolor[HTML]{7EA6E0} 0.51 & \cellcolor[HTML]{7EA6E0} 0.51 & \cellcolor[HTML]{7EA6E0} 0.51 & \cellcolor[HTML]{FF9999} 0.52 & \cellcolor[HTML]{7EA6E0} 0.50 & \cellcolor[HTML]{7EA6E0} 0.50 & 0.51 & \cellcolor[HTML]{7EA6E0} 0.50 & 0.51 \\
 \hline
\textbf{$Aver_a$} & \textbf{0.55} & \textbf{0.55} & \textbf{0.55} & \textbf{0.55} & \textbf{0.55} & \textbf{0.55} & \textbf{0.61} & \textbf{0.60} & \textbf{0.61} & \textbf{0.60} & \textbf{0.60} & \textbf{0.61} \\
 \hline
 \hline
  \textbf{$RT_g$} & \cellcolor[HTML]{FF9999} 0.64 & \cellcolor[HTML]{7EA6E0} 0.63 & \cellcolor[HTML]{7EA6E0} 0.63 & \cellcolor[HTML]{7EA6E0} 0.63 & \cellcolor[HTML]{7EA6E0} 0.63  & \cellcolor[HTML]{7EA6E0} 0.63 & \cellcolor[HTML]{7EA6E0} 0.78 & 0.79 & 0.79 & 0.79 & \cellcolor[HTML]{FF9999} 0.80 & \cellcolor[HTML]{FF9999} 0.80 \\
 \hline
 \textbf{$Sine_g$} & \cellcolor[HTML]{FF9999} 0.51 & \cellcolor[HTML]{FF9999} 0.51 & \cellcolor[HTML]{FF9999} 0.51 & \cellcolor[HTML]{FF9999} 0.51 & \cellcolor[HTML]{FF9999} 0.51 & \cellcolor[HTML]{FF9999} 0.51 & 0.51 & \cellcolor[HTML]{7EA6E0} 0.50 & 0.51 & \cellcolor[HTML]{7EA6E0} 0.50 & \cellcolor[HTML]{FF9999} 0.52 & \cellcolor[HTML]{7EA6E0} 0.50 \\
 \hline
 \textbf{$Mixed_g$} & \cellcolor[HTML]{FF9999} 0.51 & \cellcolor[HTML]{FF9999} 0.51 & \cellcolor[HTML]{FF9999} 0.51 & \cellcolor[HTML]{FF9999} 0.51 & \cellcolor[HTML]{FF9999} 0.51 & \cellcolor[HTML]{FF9999} 0.51 & \cellcolor[HTML]{7EA6E0} 0.49 & \cellcolor[HTML]{7EA6E0} 0.49 & \cellcolor[HTML]{FF9999} 0.50 & \cellcolor[HTML]{FF9999} 0.50 & \cellcolor[HTML]{7EA6E0} 0.49 & \cellcolor[HTML]{7EA6E0} 0.49 \\
 \hline
\textbf{$Aver_g$} & \textbf{0.55} & \textbf{0.55} & \textbf{0.55} & \textbf{0.55} & \textbf{0.55} & \textbf{0.55} & \textbf{0.59} & \textbf{0.59} & \textbf{0.60} & \textbf{0.60} & \textbf{0.60} & \textbf{0.60} \\
 \hline
 \hline
 \textit{ELE} & \cellcolor[HTML]{7EA6E0} 0.70 & \cellcolor[HTML]{FF9999} 0.75 & \cellcolor[HTML]{FF9999} 0.75 & \cellcolor[HTML]{FF9999} 0.75 & \cellcolor[HTML]{FF9999} 0.75 & \cellcolor[HTML]{FF9999} 0.75 & \cellcolor[HTML]{7EA6E0} 0.71 & \cellcolor[HTML]{FF9999} 0.75 & \cellcolor[HTML]{FF9999} 0.75 & \cellcolor[HTML]{FF9999} 0.75 & \cellcolor[HTML]{FF9999} 0.75 & \cellcolor[HTML]{FF9999} 0.75 \\
 \hline
 
\textit{CIC}  & \cellcolor[HTML]{7EA6E0} 0.38 & \cellcolor[HTML]{FF9999} 0.80 & \cellcolor[HTML]{FF9999} 0.80 & \cellcolor[HTML]{FF9999} 0.80 & \cellcolor[HTML]{FF9999} 0.80 & \cellcolor[HTML]{FF9999} 0.80 & \cellcolor[HTML]{7EA6E0} 0.71 & \cellcolor[HTML]{FF9999} 0.78 & \cellcolor[HTML]{FF9999} 0.78 & \cellcolor[HTML]{FF9999} 0.78 & 0.77 & 0.77 \\
 \hline
\textbf{$Aver_{r}$} & \textbf{0.54} & \textbf{0.78} & \textbf{0.78} & \textbf{0.78} & \textbf{0.78} & \textbf{0.78} & \textbf{0.71} & \textbf{0.77} & \textbf{0.77} & \textbf{0.77} & \textbf{0.76} & \textbf{0.76} \\
 \hline
\end{tabular}
    \caption{Comparison of Window-based Methods Combined With Two Base Learners}
    \label{table:Comparisons-50-window-Window}
\end{table}

\section{Comparison of Ensemble-based Detectors}

The ensemble detectors (cf. Table \ref{table:Comparisons-50-window-Ensemble}) demonstrate considerably higher performance across most dataset categories compared to SPC and window-based methods. On abrupt-drift datasets, NB combined with ARF achieves the strongest average, followed by AWE, AUE, and DWM in descending order. With HT, ARF further improves and continues to lead, with AWE, AUE, and DWM preserving the same ranking across both learners.

On gradual-drift datasets, NB with ARF again obtains the highest average, followed by AWE, AUE, and DWM. With HT, ARF continues to lead, followed by AWE, AUE, and DWM, maintaining a consistent ordering across both base learners.

On real-world data streams, the ranking shifts considerably. With NB, AUE achieves the highest average, followed by AWE, ARF, and DWM. With HT, AUE becomes the clear top performer, followed by AWE, DWM, and ARF, indicating that AUE holds a particular advantage on real-world streams while ARF's dominance does not transfer from synthetic data.

When considering all dataset types together, HT surpasses NB across all ensemble methods. ARF+HT achieves the highest overall average, closely followed by AUE+HT and AWE+HT, with DWM+HT reaching the lowest among HT-based methods. For NB, AUE leads, followed by AWE, ARF, and DWM. The results indicate that while ARF consistently dominates on both abrupt and gradual synthetic drift scenarios, AUE proves most effective on real-world data, underscoring the importance of dataset characteristics when selecting an ensemble-based drift detection method.

\begin{table}[h!]
    \small
    \centering
    \setlength{\tabcolsep}{2pt} 
    \renewcommand{\arraystretch}{1.1} 
\begin{tabular}{ |p{1.5cm}||p{1.1cm}|p{1.1cm}|p{1.1cm}|p{1.1cm}|p{1.1cm}|p{1.1cm}|p{1.1cm}|p{1.1cm}|  }
 \hline
  \multicolumn{1}{|c|}{} &
  \multicolumn{4}{|c|}{NB} &
  \multicolumn{4}{|c|}{HT} \\
 \hline
 \textbf{} & ARF & AUE & DWM & AWE & ARF & AUE & DWM & AWE \\
 \hline
 \hline
 \textbf{$RT_a$} & \cellcolor[HTML]{FF9999} 0.83 & 0.73 & \cellcolor[HTML]{7EA6E0} 0.65 & 0.76 & \cellcolor[HTML]{FF9999} 0.88 & 0.74 & \cellcolor[HTML]{7EA6E0} 0.69 & 0.77  \\
 \hline
 \textbf{$Sine_a$} & \cellcolor[HTML]{FF9999} 0.95 & 0.78 & \cellcolor[HTML]{7EA6E0} 0.59 & 0.90 & \cellcolor[HTML]{FF9999} 0.98 & 0.83 & \cellcolor[HTML]{7EA6E0} 0.80 & 0.91 \\
 \hline
 \textbf{$Mixed_a$} & \cellcolor[HTML]{FF9999} 0.97 & 0.68 & \cellcolor[HTML]{7EA6E0} 0.56 & 0.87 & \cellcolor[HTML]{FF9999} 0.97 & 0.78 & \cellcolor[HTML]{7EA6E0} 0.68 & 0.89 \\
 \hline
\textbf{$Aver_a$} & \textbf{0.92} & \textbf{0.73} & \textbf{0.60} & \textbf{0.84} & \textbf{0.94} & \textbf{0.78} & \textbf{0.72} & \textbf{0.86} \\
 \hline
 \hline
  \textbf{$RT_g$} & 0.76 & 0.73 & \cellcolor[HTML]{7EA6E0} 0.65 & \cellcolor[HTML]{FF9999} 0.78 & \cellcolor[HTML]{FF9999} 0.88 & \cellcolor[HTML]{7EA6E0} 0.74 & 0.76 & 0.80 \\
 \hline
 \textbf{$Sine_g$} & \cellcolor[HTML]{FF9999} 0.93 & 0.81 & \cellcolor[HTML]{7EA6E0} 0.53 & 0.88 & \cellcolor[HTML]{FF9999} 0.97 & 0.84 & \cellcolor[HTML]{7EA6E0} 0.53 & 0.90 \\
 \hline
 \textbf{$Mixed_g$} & \cellcolor[HTML]{FF9999} 0.96 & 0.70 & \cellcolor[HTML]{7EA6E0} 0.53 & 0.84 & \cellcolor[HTML]{FF9999} 0.96 & 0.79 & \cellcolor[HTML]{7EA6E0} 0.54 & 0.87 \\
 \hline
\textbf{$Aver_g$} & \textbf{0.88} & \textbf{0.75} & \textbf{0.57} & \textbf{0.83} & \textbf{0.94} & \textbf{0.72} & \textbf{0.61} & \textbf{0.86} \\
 \hline
 \hline
 \textit{ELE} & 0.81 & \cellcolor[HTML]{FF9999} 0.88 & \cellcolor[HTML]{7EA6E0} 0.70 & 0.74 & 0.82 & \cellcolor[HTML]{FF9999} 0.89 & \cellcolor[HTML]{7EA6E0} 0.71 & 0.75 \\
 \hline
\textit{CIC} & \cellcolor[HTML]{7EA6E0} 0.36 & \cellcolor[HTML]{FF9999} 0.70 & 0.37 & 0.64 & \cellcolor[HTML]{7EA6E0} 0.63 & \cellcolor[HTML]{FF9999} 0.87 & 0.77 & 0.75 \\
 \hline
\textbf{$Aver_{r}$} & \textbf{0.59} & \textbf{0.79} & \textbf{0.54} & \textbf{0.69} & \textbf{0.73} & \textbf{0.88} & \textbf{0.74} & \textbf{0.75} \\
 \hline
\end{tabular}
    \caption{Comparisons of Ensemble-based Methods Combined With Two Base Learners}
    \label{table:Comparisons-50-window-Ensemble}
\end{table}

\section{A Summary}
As presented in Table \ref{tab:best_methods}, HT generally outperforms NB across most dataset categories and detector families. However, a notable exception arises on real-world data streams, where NB achieves equal or slightly superior performance to HT within both SPC and window-based methods. This suggests that the advantage of a more expressive base learner is not universal and depends on the nature of the data. Ensemble methods consistently outperform both SPC and window-based detectors across all dataset types, and their strongest results are achieved with HT, reinforcing that the choice of base learner significantly impacts predictive accuracy under concept drift.

Focusing on single-detector families, EWMA+HT and EDDM+HT offer the best overall SPC performance ($\approx 0.69$) and the best gradual-drift accuracy ($\approx 0.60$). For real-world streams, NB paired with RDDM, FHDDM, EWMA, or EDDM achieves the highest SPC result of approximately $0.78$. Within the window-based family, KSWIN+HT, WSTD+HT, and D3+HT are jointly strongest for abrupt drifts ($\approx 0.61$), while WSTD+HT, MDDM+HT, ADWIN+HT and D3+HT perform best on gradual drifts ($\approx 0.60$). On real-world streams, all window-based methods paired with NB uniformly achieve the top result of approximately $0.78$. Overall, all window-based methods with HT yield an identical aggregate performance of $0.68$, with no single detector emerging as dominant. Among ensemble-based methods, ARF+HT achieves the strongest results on both abrupt and gradual drift scenarios ($\approx 0.94$) and ranks best in overall aggregate performance ($\approx 0.83$), while AUE+HT proves most effective on real-world data streams ($\approx 0.88$). Overall, although SPC and window-based detectors remain competitive, particularly on real-world data with NB, the most reliable and consistent improvements across all drift scenarios come from ensemble-based adaptation mechanisms, especially ARF+HT.

\begin{table}[t]
\centering
\label{tab:best_methods}
\begin{tabular}{llll}
\hline
\textbf{Cat.} & \textbf{Drift type} & \textbf{Learner} & \textbf{Best method} \\
\hline
SPC   & Abrupt   & HT  & FTDD  \\
SPC   & Gradual  & HT  & EWMA/EDDM  \\
SPC   & Real-world  & NB  & RDDM/FHDDM/EWMA/EDDM  \\
   & Overall    & HT  & EWMA/EDDM  \\
\hline
Window & Abrupt   & HT  & KSWIN/WSTD/D3 \\
Window & Gradual  & HT  & WSTD/MDDM/ADWIN/D3  \\
Window & Real-world  & NB & FPDD/WSTD/MDDM/ADWIN/D3  \\
 & Overall     & HT & WSTD/D3  \\
\hline
Ensemble     & Abrupt   & HT & ARF  \\
Ensemble     & Gradual  & HT & ARF  \\
Ensemble     & Real-world  & HT & AUE \\
     & Overall  & HT  & ARF  \\
\hline
\end{tabular}
\caption{Best-performing drift detectors by category, drift type and base learner}
\end{table}

\section{Conclusions}
Concept drifts are prevalent in today's streaming applications where data distributions and relationships change over time. They have significant consequences in many critical domains,  such as healthcare, banking, finance, cybersecurity, email spam and phishing, IoT, sensor networks and recommendation systems. In these applications, drifts can degrade the performance of decision models and increase the risk of incorrect classifications. Therefore, accurately identifying such drifts is crucial to prevent classification models from deteriorating.\\ 
 
The aim of our study is to simplify the description of a broad range of learner-based concept drift detection algorithms in order to enhance the comprehension of these complex methods.  These algorithms are classified into three main groups, Statistical Process Control (SPC) methods, windowing techniques and ensemble-based approaches.  These methods can operate in either active or passive modes and can be implemented in online or block-based settings. We have also conducted an extensive evaluation and comparison of these detectors using both synthetic and real-world datasets to assess their performance.

\printbibliography
\end{document}